\documentclass[letterpaper, 10 pt, journal, twoside]{ieeetran}
\usepackage{times}

\usepackage[numbers]{natbib}
\usepackage{multicol}
\usepackage[bookmarks=true]{hyperref}
\usepackage{amsfonts} 
\usepackage{xcolor}
\usepackage{graphicx}
\usepackage{amsmath}
\usepackage{booktabs} 
\usepackage{colortbl}
\usepackage{tabularray}
\usepackage{multirow}
\usepackage{fontawesome5}

\DeclareMathOperator*{\argmin}{arg\,min}
\DeclareMathOperator*{\argmax}{arg\,max}

\definecolor{green}{RGB}{11,155,13}

\begin{document}

\title{\LARGE CAHSOR: Competence-Aware High-Speed Off-Road\\ Ground Navigation in $\mathbb{SE}(3)$}

\author{Anuj Pokhrel, Mohammad Nazeri, Aniket Datar, and Xuesu Xiao\\
\thanks{All authors are with the Department of Computer Science, George Mason University
        {\tt\footnotesize \{apokhre, mnazerir, adatar, xiao\}@gmu.edu}}}




%

\maketitle
\thispagestyle{empty}
\pagestyle{empty}

\begin{abstract}
While the workspace of traditional ground vehicles is usually assumed to be in a 2D plane, i.e., $\mathbb{SE}(2)$, such an assumption may not hold when they drive at high speeds on unstructured off-road terrain: High-speed sharp turns on high-friction surfaces may lead to vehicle rollover; Turning aggressively on loose gravel or grass may violate the non-holonomic constraint and cause significant lateral sliding; Driving quickly on rugged terrain will produce extensive vibration along the vertical axis. Therefore, most offroad vehicles are currently limited to drive only at low speeds to assure vehicle stability and safety. In this work, we aim at empowering high-speed off-road vehicles with competence awareness in $\mathbb{SE}(3)$ so that they can reason about the consequences of taking aggressive maneuvers on different terrain with a 6-DoF forward kinodynamic model. The kinodynamic model is learned from visual and inertial Terrain Representation for Off-road Navigation (\textsc{tron}) using multimodal, self-supervised vehicle-terrain interactions. We demonstrate the efficacy of our Competence-Aware High-Speed Off-Road (\textsc{cahsor}) navigation approach on a physical ground robot in both an autonomous navigation and a human shared-control setup and show that \textsc{cahsor} can efficiently reduce vehicle instability by 62\% while only compromising 8.6\% average speed with the help of \textsc{tron}\footnote{\faIcon{github}~\url{https://github.com/AnujPokhrel/CAHSOR}}.
\end{abstract}


\IEEEpeerreviewmaketitle

\section{INTRODUCTION}
\label{sec::intro}
Autonomous mobile robot navigation has been a research topic in the robotics community for decades~\cite{fox1997dynamic, quinlan1993elastic}. Being equipped with perception, planning, and control techniques, different types of ground robots, e.g., differential-drive or Ackermann-steering, can efficiently move toward their goals in their 2D workspaces considering their 3-DoF motion models ($x$, $y$, and $\textrm{yaw}$) without colliding with obstacles, mostly in structured and homogeneous environments~\cite{xiao2022autonomous, xiao2023autonomous}. 

\begin{figure}
  \centering
  \includegraphics[width=\columnwidth]{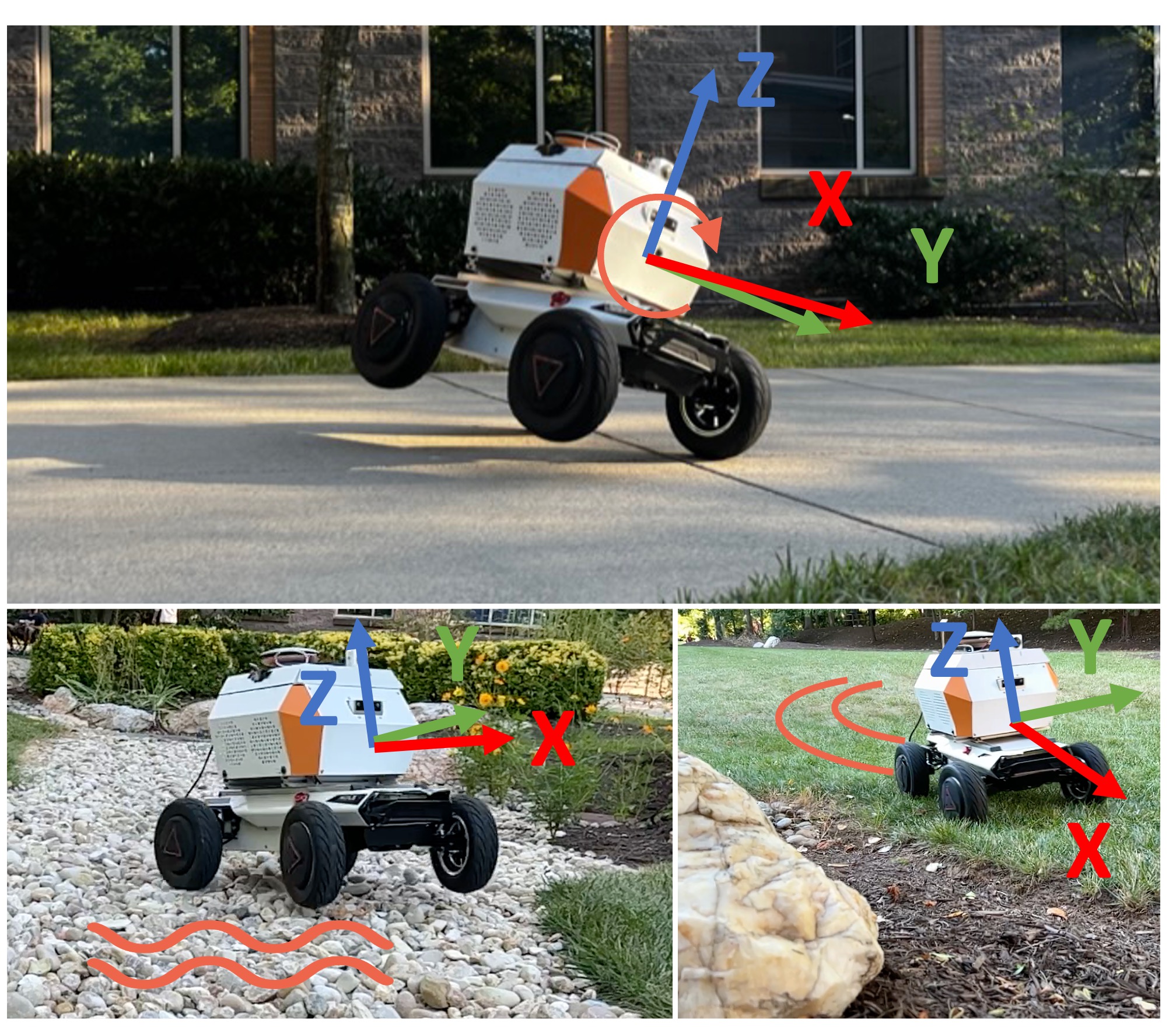}
  \caption{Challenges of High-Speed Off-Road Ground Navigation in $\mathbb{SE}(3)$.}
  \label{fig::hunter}
  \vspace{-9pt}
\end{figure}

Bringing those robots into the unstructured real world, researchers have also investigated off-road navigation since the DARPA Grand Challenge~\cite{seetharaman2006unmanned} and LAGR (Learning Applied to Ground Vehicles) Program~\cite{jackel2006darpa}. While significant research effort on off-road navigation focuses on the perception side~\cite{meng2023terrainnet, wolf2020advanced, bai2019three, maturana2018real}, researchers have also investigated off-road mobility, including inverse~\cite{karnan2022vi, xiao2021learning} and forward~\cite{atreya2022high, maheshwari2023piaug} kinodynamics,  wheel slip modeling~\cite{rogers2012aiding, rabiee2019friction}, and end-to-end learning~\cite{pan2020imitation, siva2019robot}. Most off-road robots drive at slow speeds to assure vehicle stability and safety~\cite{datar2023toward, datar2023learning}. Even when aiming at driving fast, they still assume a simplified 2D workspace and 3-DoF model in $\mathbb{SE}(2)$ despite the highly likely disturbances from the off-road terrain on other dimensions of the state space (e.g., drift along $y$, roll around $x$, or bumpiness along $z$). These realistic kinodynamic effects may be tolerable in some cases, but may lead to catastrophic consequences in others with increasing speed on unstructured terrain (Fig.~\ref{fig::hunter}). 

To enable safe and robust off-road navigation, high-speed ground robots need to be competence-aware, i.e., knowing what is the consequence of taking an aggressive maneuver on different off-road terrain. For example, a sharp turn on high-friction pavement may lead to vehicle rollover (Fig.~\ref{fig::hunter} top); Moving at high speeds through rugged surfaces can generate extensive vertical vibrations and damage onboard components (Fig.~\ref{fig::hunter} bottom left); Aggressive swerving on loose grass or gravel will cause the vehicle to slide sideways and risk collision or falling off a cliff (Fig.~\ref{fig::hunter} bottom right). 

To this end, we propose a Competence-Aware High-Speed Off-Road (\textsc{cahsor}) ground navigation approach based on a 6-DoF forward kinodynamic model in $\mathbb{SE}(3)$. The model is learned as a downstream task of a new Terrain Representation for Off-road Navigation (\textsc{tron}) approach with multimodal, self-supervised learning using viewpoint-invariant visual terrain patches complemented with vehicle speed and underlying Inertia Measurement Unit (IMU) responses during vehicle-terrain interactions. \textsc{cahsor} learns to predict potential next states according to different candidate actions and the current vision-speed and/or inertia terrain representation to make competence-aware decisions in order to maximize speed while satisfying 6-DoF vehicle stability constraints in $\mathbb{SE}(3)$,
e.g., without excessive sliding and rolling motions or bumpy vibrations.  
Our contributions can be summarized as: 
\begin{itemize}
    \item a \textsc{tron} approach with multimodal self supervision that allows onboard vision-speed and inertia observations to augment each other and maximizes the information embedded in the representation of each perceptual modality;
    \item a comprehensive study of various end-to-end and representation learning techniques with different modalities for different off-road kinodynamic modeling tasks;
    \item a \textsc{cahsor} framework for high-speed off-road vehicles to take aggressive maneuvers with stability and safety;
    \item a set of real-world, off-road robot experiments to demonstrate the effectiveness of \textsc{cahsor} based on \textsc{tron} in both autonomous navigation and a human shared-control setup, exhibiting 62\% vehicle instability reduction while only compromising 8.6\% average speed.
\end{itemize}

\section{RELATED WORK}
\label{sec::related}

We review related robot navigation research focusing on off-road conditions and using machine learning approaches. 

\subsection{Off-Road Navigation}
Since the DARPA Urban Challenge~\cite{seetharaman2006unmanned} and LAGR Program~\cite{jackel2006darpa}, robotics researchers have investigated autonomous navigation techniques for off-road conditions. 
Robot perception is the first challenge in extending the basic distinction between obstacles and free spaces, e.g., identifying semantic information such as pavement, gravel, grass, pebbles, and mud. 
Terrain classification methods perceive the underlying terrain and make navigation decisions tailored to the terrain class~\cite{bai2019three, cai2022risk, cai2023probabilistic}, 
while terrain segmentation approaches use terrain semantics and build traversability costmaps to inform subsequent planning~\cite{maturana2018real, wolf2020advanced, viswanath2021offseg, triest2022tartandrive, meng2023terrainnet, karnan2023sterling, dixit2023step, howdoesitfeel}. 
Furthermore, off-road navigation has been investigated by robotics researchers from the mobility side, including vehicle dynamics modeling~\cite{triest2022tartandrive, karnan2022vi, xiao2021learning} and drifting~\cite{hindiyeh2013dynamics}. 
While most off-road robots are treated as vehicles moving in a 2D workspace with 3 DoFs ($x$, $y$, and $\textrm{yaw}$), recent work has investigated the effect of uneven terrain on vehicle $\mathbb{SE}(3)$ state (with the additional $z$, $\textrm{pitch}$, and $\textrm{roll}$)~\cite{datar2023toward, datar2023learning, han2023model}.
Our \textsc{cahsor} also operates in $\mathbb{SE}(3)$, but our 6-DoF forward kinodynamic model aims at confidently navigating ground robots at the maximum possible speed on various semantically different, relatively flat off-road terrain while maintaining stability on other state dimensions, e.g., drift along $y$, roll around $x$, or bumpiness along $z$, which are often ignored by traditional 2D models.


\subsection{Machine Learning for Navigation}
Recently, machine learning approaches have been widely adopted to enable autonomous mobile robot navigation~\cite{xiao2022motion}. These methods enable navigation behaviors in a data-driven manner without designing systems and components which often fail to capture real-world complexities. In particular, researchers have used machine learning to learn end-to-end systems~\cite{karnan2022voila, pan2020imitation, pfeiffer2017perception, chen2017decentralized}, local planners~\cite{xu2023benchmarking, liu2021lifelong, francis2020long, wang2021agile, xiao2021agile, xiao2021toward}, planner parameterization~\cite{xiao2020appld, wang2021appli, wang2021apple, xu2021applr, xiao2022appl, xu2021machine}, kinodynamic models~\cite{lee2023learning, karnan2022vi, xiao2021learning, atreya2022high, nagariya2020iterative}, and cost functions~\cite{xiao2022learning, sikand2022visual, kretzschmar2016socially, kim2016socially, ruetz2023foresttrav, dashora2022hybrid}, for tasks like highly-constrained~\cite{xiao2022autonomous, xiao2023autonomous, wang2021agile, xiao2021agile, xiao2021toward, perille2020benchmarking, nair2022dynabarn}, social~\cite{park2023learning, xiao2022learning, karnan2022socially, nguyen2023toward, francis2023principles, kretzschmar2016socially, chen2017decentralized, kim2016socially, hart2020using, mirsky2021conflict}, and off-road navigation~\cite{pan2020imitation, karnan2022vi, xiao2021learning, atreya2022high}. Considering the complexity of unstructured off-road terrain and their intricate effects on vehicle kinodynamics at high speeds, we also adopt a data-driven method for \textsc{cahsor} to learn a forward kinodynamic function conditioned on viewpoint-invariant vision-speed and underlying inertia terrain representation learned by \textsc{tron} using multimodal, self-supervised vehicle-terrain interaction experiences. 

Comparing to existing classical or learning-based off-road navigation approaches, \textsc{cahsor} extends vehicle kinodynamics from $\mathbb{SE}(2)$~\cite{xiao2021learning, karnan2022vi, atreya2022high, pan2020imitation} to $\mathbb{SE}(3)$ and considers not only terrain geometric information~\cite{lee2023learning, datar2023learning, datar2023toward}, but also multimodal perception input from vision, inertia, and vehicle speed to reason about the consequences of aggressive maneuvers affected by a variety of environmental factors. Furthermore, in contrast to the large body of literature on off-road perception~\cite{meng2023terrainnet, wolf2020advanced, maturana2018real, karnan2023sterling, dixit2023step, viswanath2021offseg, triest2022tartandrive} to decide \emph{where to drive}  using a semantic cost function, \textsc{cahsor} provides a solution to \emph{how to drive} at the maximum possible speed without incurring catastrophic consequences given different underlying terrain. To the best of our knowledge, such high-speed competence awareness by \textsc{cahsor} is not achievable by any state-of-the-art off-road representation~\cite{karnan2023sterling} and kinodynamics~\cite{lee2023learning} learning approaches (see experiment results in Sec.~\ref{sec::experiments}).
\section{APPROACH}
\label{sec::approach}
We formulate the problem of forward kinodynamic modeling in $\mathbb{SE}(3)$ for ground robots driving on unstructured off-road terrain at high speeds. Our approach employs a multimodal self-supervised learning approach to represent off-road conditions using onboard visual, inertial, and speed observations. Additionally, we introduce a data-driven approach to learn the forward kinodynamic model from past vehicle-terrain interactions. Finally, we develop a competence-aware navigation framework that allows robots to drive at the maximum possible speed while maintaining vehicle stability in $\mathbb{SE}(3)$. 

\begin{figure}
  \centering
  \includegraphics[width=\columnwidth]{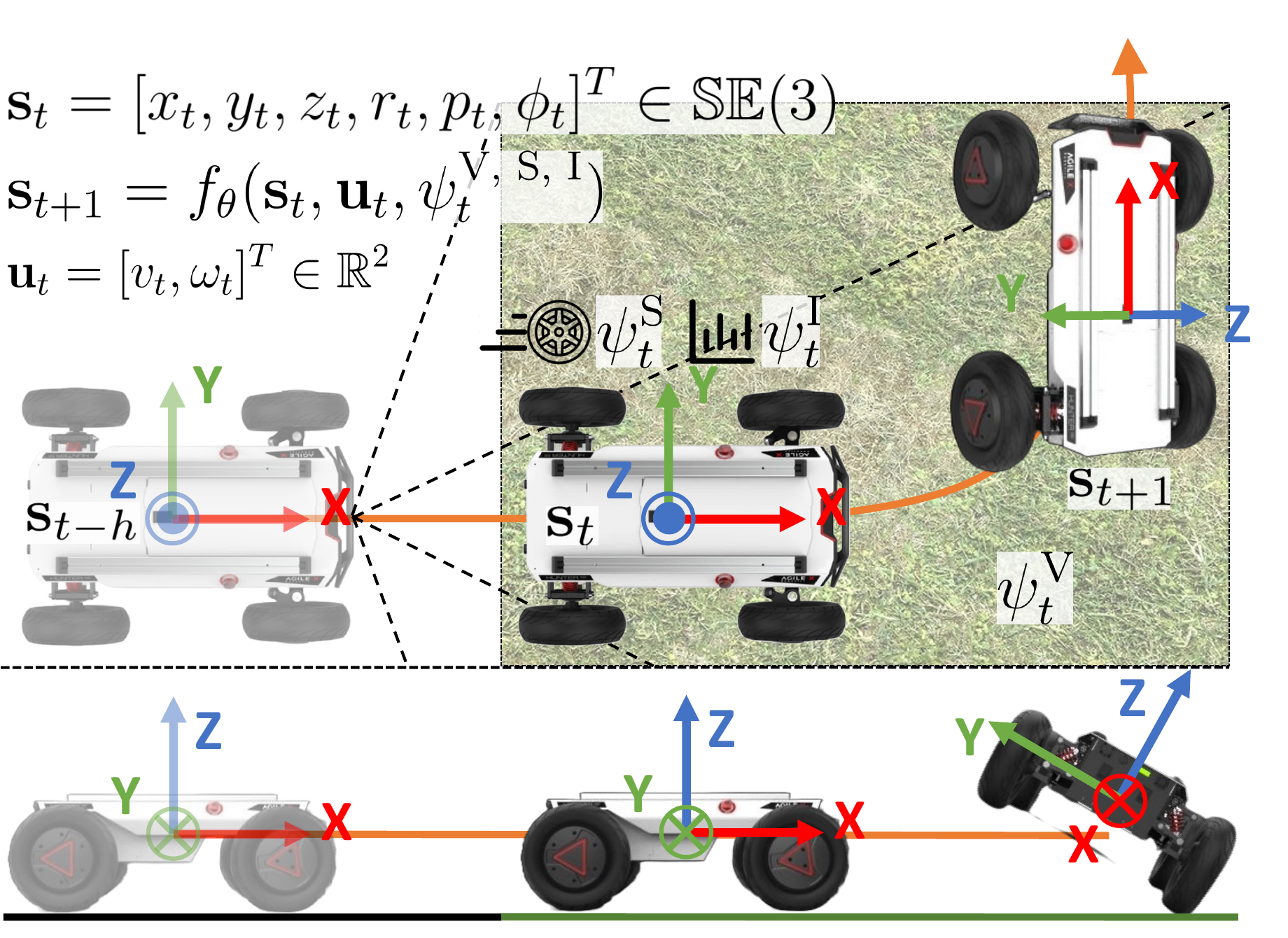}
  \caption{Overview of the Competence-Aware High-Speed Off-Road (\textsc{cahsor}) Ground Navigation (Side and Top View) based on Vision ($\psi^{V}$), Inertia ($\psi^I$), and Speed($\psi^{S}$) Representation.}
  \label{fig::cahsor}
\end{figure}

\subsection{Forward Ground Kinodynamics in $\mathbb{SE}(3)$}
We adopt a forward kinodynamics formulation for ground robots to reason about the consequences of taking different aggressive maneuvers on various off-road terrain. We denote vehicle state as $\mathbf{s}$, which includes 6-DoF vehicle pose in $\mathbb{SE}(3)$ ($x$, $y$, $z$, $\textrm{roll}$ $r$, $\textrm{pitch}$ $p$, and $\textrm{yaw}$ $\phi$, expressed in the global or robot frame, Fig~\ref{fig::cahsor}) and their corresponding velocity components. For brevity, only the pose components are included in the following derivation. The vehicle control $\mathbf{u} = [v, \omega]^T$ contains linear velocity and angular velocity (for differential-driven vehicles, or steering curvature for Ackermann-steering vehicles). We use a world state $\mathbf{w}$ to denote all necessary environmental effects on kinodynamics, in our case, from unstructured off-road terrain. Therefore, in a discrete setting, we have 
\begin{equation}
\begin{gathered}
   \mathbf{s}_{t+1}=f(\mathbf{s}_t, \mathbf{u}_t, \mathbf{w}_t),\quad \mathbf{o}_t = g(\mathbf{s}_t, \mathbf{w}_t), \\
   \mathbf{s}_t= [x_t, y_t, z_t, r_t, p_t, \phi_t]^T \in \mathbb{SE}(3),~\mathbf{u}_t=[v_t, \omega_t]^T \in \mathbb{R}^2,
\end{gathered}
\nonumber
\end{equation}
where $f(\cdot)$ is a forward kinodynamic function in $\mathbb{SE}(3)$, while $g(\cdot)$ is an observation function. For off-road driving, the forward kinodynamic function $f(\cdot)$ also takes in the world state $\mathbf{w}$ as input, in contrast to models derived for structured and homogeneous terrain that only need to consider vehicle state $\mathbf{s}_t$ and control $\mathbf{u}_t$ alone. For example, slippery and rugged terrain surfaces may cause extensive movement along the vehicle $y$ and $z$ directions respectively. However, world state $\mathbf{w}$ is not directly observable or easily modeled. Also notice that most kinodynamic models for ground robots only consider $\mathbf{s}_t = [x_t, y_t, \phi_t]^T \in \mathbb{SE}(2)$ and ignore all other state dimensions. Such a forward model can be used in rolling out candidate trajectories for sampling-based path and motion planners~\cite{williams2017model, fox1997dynamic}, or its inverse form can be derived to achieve desired next state $\mathbf{s}^*_{t+1}$~\cite{xiao2021learning, karnan2022vi}. 

\subsection{Visual and Inertial Representation of World State}
Considering the difficulty in analytically modeling world state $\mathbf{w}$, we use a multimodal self-supervised learning approach to represent $\mathbf{w}$ and approximate the observation function $g(\cdot)$ with onboard visual and inertial sensors. 
In particular, an onboard visual camera can provide a visual signature of the terrain patch $\lambda_t$ the robot drives over and an IMU can sense the underlying kinodynamic responses $i_t$ in terms of linear accelerations and angular velocities. 
We assume the current vehicle speed can also be observed by, e.g., odometry or GPS, denoted as $s_t$. 
While IMU readings $i_t$ can be directly sensed when in contact with the underlying terrain, for high-speed off-road navigation, the vehicle may need to reason about future kinodynamic consequences up to a certain planning horizon, for which only visual observations from a forward-looking camera are available, not the IMU readings. 
So in this work, we use multimodal self-supervised learning to allow both visual and inertial observations to augment each other by correlating them in effective representation spaces, thus either (or both) can be used to enable competence awareness when available (e.g., manual shared-control using current underlying inertia and autonomous planning with the vision of future terrain). 

We posit that the visual and inertial observations can provide multimodal self-supervised learning signals to represent different terrain kinodynamics. To achieve such self-supervision, we use a non-contrastive approach to maximize the correlation between visual and inertial embeddings. In this way, we avoid the need to use privileged information such as terrain labels that require manual annotation.
However, a key difference of high-speed off-road navigation compared to existing terrain representation learning approaches~\cite{sikand2022visual, karnan2023sterling} is that  the correlation between vision and inertia is also dependent on the (high) vehicle speed: driving quickly vs. slowly on the same visual patch of grass may produce completely different inertial responses, while different speeds on grass vs. gravel may coincidentally lead to similar IMU readings. Therefore, \textsc{cahsor} extends the vision--inertia correlation to vision \& speed--inertia correlation to account for the effect caused by various speeds during high-speed off-road navigation. 

However, two challenges still exist for visual representation: 
1. Visual perception is very sensitive to the environment, such as changes in viewpoints and lighting, as well as occlusion and motion blur; 
2. Unlike current IMU readings, the visual signature of the terrain underneath (or right in front of) the current robot state $\mathbf{s}_t$ can not be directly captured by the onboard camera due to limited onboard field-of-view. The robot needs to seek help from previous camera images captured before $t$. 
Therefore, we design a viewpoint-invariant visual patch extraction technique to overcome both challenges: 
Denote the camera image captured $h$ time steps before as $c_{t-h}$ and the transformation from time step $t-h$ to $t$ extracted from vehicle odometry as $d^t_{t-h}$. By projecting $c_{t-h}$ to an overhead Bird-Eye View (BEV) using the camera homography $h_{t-h}=H(i_{t-h})$, which is dependent on the vehicle roll and pitch angles determined by the IMU readings $i_{t-h}$ due to aggressive off-road driving, we can extract the terrain patch currently underneath the robot, $\lambda_t = P(c_{t-h}, h_{t-h}, d^t_{t-h})$. $\lambda_t$ is designed to be slightly larger than the vehicle footprint to consider actuation latency. By varying the history length $h$, it is possible to generate a set of different instantiations of $\lambda_t$ from different viewpoints with different lighting conditions, $\Lambda_t=\{\lambda_t^j\}_{j=1}^{J}$, in which each $\lambda_t^j$ has at least a certain amount of visible pixels of the terrain patch underneath $\mathbf{s}_t$, considering the homography projection may cause invisible BEV pixels. 

\begin{figure*}
  \centering
  \includegraphics[width=2\columnwidth]{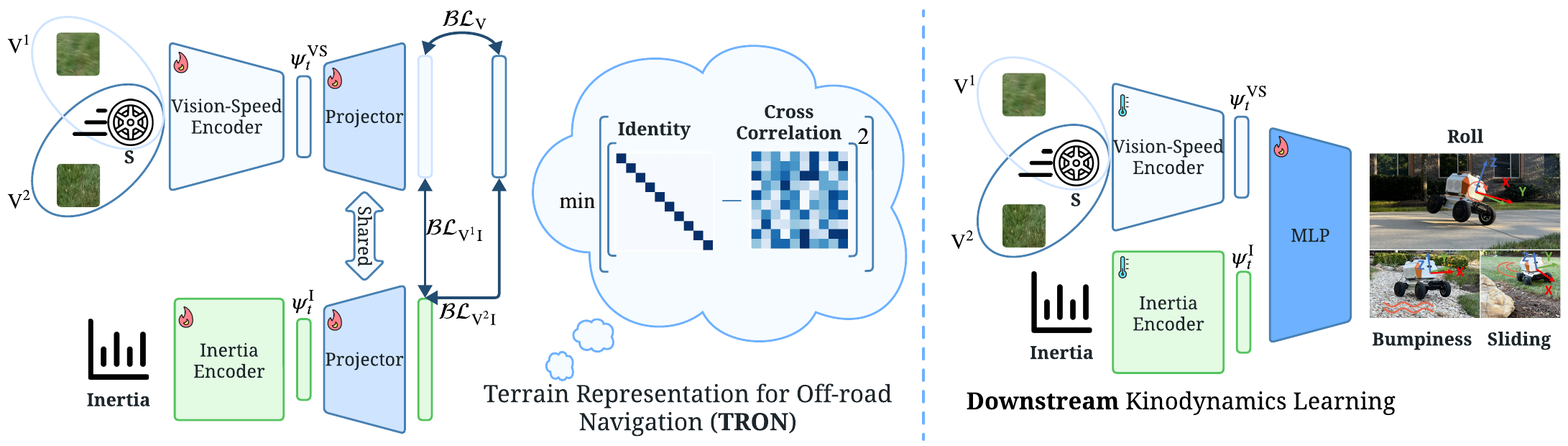}
  \caption{\textsc{tron} (Left) and Downstream Kinodynamics Learning (Right) Architecture: Flame and temperature denote training and frozen parameters respectively.}
  \label{fig::nn}
\end{figure*}

\subsection{Terrain Representation for Off-road Navigation}
The set of viewpoint-invariant visual terrain patches $\Lambda_t=\{\lambda_t^j\}_{j=1}^{J}$, the IMU readings $i_t$, and the current vehicle speed $s_t$ correspond to multimodal perception of the robot at time $t$ and provide self-supervised learning signals for our vision \& speed--inertia correlation (Fig.~\ref{fig::nn} left). To be specific, a vision, speed, and inertia encoder embeds any terrain patch $\lambda_t \in \Lambda_t$, current vehicle speed $s_t$, and underlying inertial readings $i_t$ into a visual, speed, and inertial representation, $\psi^\textrm{V}_t$, $\psi^\textrm{S}_t$, and $\psi^\textrm{I}_t$, respectively. Considering the causal relation from driving at a particular speed on a certain visual terrain patch to corresponding IMU readings, we concatenate the visual and speed representations, $\psi^\textrm{V}_t$ and $\psi^\textrm{S}_t$, and further encode them into a joint vision \& speed embedding $\psi^\textrm{VS}_t$. To correlate $\psi^\textrm{VS}_t$ and $\psi^{I}_t$, we project them independently into a higher dimensional feature space, $\rho^\textrm{VS}_t$ and $\rho^{I}_t$. We then maximize the correlation between $\rho^\textrm{VS}_t$ and $\rho^{I}_t$ while considering viewpoint invariance using Barlow Twins~\cite{zbontar2021barlow}:
\begin{equation}
\begin{split}
    \mathcal{L}_\textrm{TRON} = \mathcal{BL}_\textrm{V}(\rho^{\textrm{V}^1\textrm{S}}_t, \rho^{\textrm{V}^2\textrm{S}}_t) &+ \\
    ( 0.5 \times \mathcal{BL}_{\textrm{V}^1\textrm{I}}(\rho^{\textrm{V}^1\textrm{S}}_t, \rho^{I}_t) &+ 
    0.5 \times \mathcal{BL}_{\textrm{V}^2\textrm{I}}(\rho^{\textrm{V}^2\textrm{S}}_t, \rho^{I}_t)),
    \label{eqn::cahsor_ssl}
\end{split}
\end{equation}
where $\textrm{V}^1$ and $\textrm{V}^2$ correspond to two views of the same terrain patch to encourage viewpoint invariance, i.e., $\lambda^1_t, \lambda^2_t \sim \Lambda_t$. $\mathcal{BL}$ is defined as:
\begin{equation}
    \mathcal{BL}= \sum_i(1-\mathcal{C}_{ii})^2 + \gamma \sum_i \sum_{j \neq i} \mathcal{C}_{ij}^2,
    \label{eqn::barlow}
\end{equation}
where $\gamma$ is a weight term to trade off the importance between invariance and redundancy reduction. $\mathcal{C}$ is the cross-correlation matrix computed between $\rho^1$ and $\rho^2$:
\begin{equation}
    \mathcal{C}_{ij} = \frac{\sum_b \rho^1_{b, i}\rho^2_{b, j}}{\sqrt{\sum_b (\rho^1_{b, i})^2}\sqrt{\sum_b (\rho^2_{b, j})^2}},
    \nonumber
\end{equation}
where $\rho^{1(2)}_{b, i}$ denotes the $i$th dimension of the $b$th sample in a data batch of $\rho^{1(2)}$, which can be one of $\rho^{\textrm{V}^1\textrm{S}}_t$, $\rho^{\textrm{V}^2\textrm{S}}_t$, or $\rho^{I}_t$. 

Trained with multimodal self-supervision, the visual, speed, and inertial representation, $\psi^\textrm{V}_t$, $\psi^\textrm{S}_t$, and $\psi^\textrm{I}_t$, can be used to enable downstream kinodynamic modeling tasks. Depending on the scenario, either $\psi^\textrm{V}_t$ (predicting multiple future states without terrain interactions to induce inertial responses) or $\psi^\textrm{I}_t$ (directly predicting the immediate next state from the induced inertial responses from the underlying terrain), or both, may be available. For simplicity, we denote our visual-speed, inertial, or visual-speed-inertial (by concatenating the first two) representation as $\psi^\textrm{V, S, I}_t$. 

\subsection{Downstream Kinodynamic Model Learning}
After learning the terrain representation and freezing the learned parameters, we also adopt a self-supervised approach to learn the forward kinodynamics due to the difficulty in analytically modeling $f(\cdot)$. We represent the unknown world state $\mathbf{w}_t$ using $\psi^\textrm{V, S, I}_t$ and learn an approximate forward kinodynamic function $f_\theta(\cdot)$ as a downstream task of \textsc{tron}: 
\begin{equation}
    \mathbf{s}_{t+1} = f_\theta(\mathbf{s}_t, \mathbf{u}_t, \psi^\textrm{V, S, I}_t). 
    \label{eqn::approximated_fkd}
\end{equation}
With a self-supervised vehicle-terrain interaction dataset, 
\[
\mathcal{D} = \{\mathbf{s}_{j+1}, \mathbf{s}_j, \mathbf{u}_j, \psi^\textrm{V, S, I}_j\}_{j=0}^{N-1},
\]
of $N$ data points, the optimal parameters $\theta^*$ can then be learned by minimizing a supervised loss function (Fig.~\ref{fig::nn} right), 
\begin{equation}
    \theta^*=\argmin_\theta \mathop{\sum_{\substack{(\mathbf{s}_{j+1}, \mathbf{s}_j, \mathbf{u}_j, \psi^\textrm{V, S, I}_j)\\\in \mathcal{D}}}} ||\mathbf{s}_{j+1}-f_\theta(\mathbf{s}_j, \mathbf{u}_j, \psi^\textrm{V, S, I}_j)||.
    \label{eqn::supervised_loss}
\end{equation}

\subsection{Competence-Aware High-Speed Off-Road Navigation}
The approximate forward kinodynamic function (Eqn.~(\ref{eqn::approximated_fkd})) learned with the self-supervised loss (Eqn.~(\ref{eqn::supervised_loss})) can be combined with subsequent planners, e.g., sampling-based model predictive motion planners~\cite{williams2017model, fox1997dynamic}, or used in human shared-control settings  to enable competence-aware off-road navigation at high speeds. By rolling out the forward kinodynamic model, the robot can pick the optimal control command(s) that produces the minimal cost or is most similar to human control, without violating vehicle stability constraints. While a motion planner or a human controller needs to consider a variety of costs including obstacle avoidance, goal distance, execution accuracy, etc., for simplicity, we combine all these costs into one general cost term $C(\mathbf{s}_t, \mathbf{s}_{t+1})$ and use only one time-step rollout in our presentation to explicitly showcase the high speed and competence awareness aspect of the navigation problem. Notice that it is easy to combine it with any other costs when necessary and extend to multiple time steps (see examples in Sec.~\ref{sec::experiments}). Expressing the robot $\mathbb{SE}(3)$ state in the current robot frame (i.e., $x$ forward, $y$ left, and $z$ up), the competence-aware navigation can be formulated as a constrained optimization problem: 
\begin{equation}
    \begin{gathered}
        \mathbf{u}^*_t = \argmax_{\mathbf{u}_t} \left[ ||x_{t+1}-x_t|| - C(\mathbf{s}_t, \mathbf{s}_{t+1})\right], \\
        \textrm{s.t.}~\textrm{all $\mathbb{SE}(3)$ constraints are satisfied}, \\
        \mathbf{s}_{t+1} = f_\theta(\mathbf{s}_t, \mathbf{u}_t, \psi^\textrm{V, S, I}_t).
    \end{gathered}
    \label{eqn::optimization}
\end{equation}
Notice that the objective function in Eqn.~\eqref{eqn::optimization} can be formulated in other ways when necessary. 
For example, maximizing the displacement along $x$ can be replaced by minimizing the difference between the control $u$ and a desired manual command (see Sec.~\ref{sec::implementations} for two possible instantiations of Eqn.~\eqref{eqn::optimization}). 
The navigation planner then finds the best control $\mathbf{u}^*_t$ to maximize speed along $x$ (and considers other costs in $C(\cdot, \cdot)$), while in a human shared-control setup $\mathbf{u}^*_t$ aims to minimize the difference compared to human command, both without violating $\mathbb{SE}(3)$ vehicle kinodynamic constraints.

\section{IMPLEMENTATIONS}
\label{sec::implementations}

We implement \textsc{cahsor} on a 1/6-scale autonomous vehicle, an AgileX Hunter SE, with a top speed of 4.8m/s on different off-road terrain at high speeds to demonstrate the proposed competence awareness. We collect a dataset of 30-minute vehicle-terrain interactions. The collected GPS-RTK, onboard IMU, front-facing OAK-D Pro camera, and vehicle control data are synchronized and processed into training data. We integrate the learned \textsc{tron} and downstream kinodynamic models and the \textsc{cahsor} framework with an autonomous navigation planner and a human shared-control setup. 

\subsection{\textsc{cahsor} Implementations}
\subsubsection{\textsc{tron}}
The terrain vision encoder is a 4-layer Convolutional Neural Network (CNN) to produce a 512-dimensional viewpoint-invariant visual representation.
The speed encoder is a 2-layer neural network, whose 512-dimensional output is combined with the visual representation to construct our vision-speed representation $\psi^\textrm{VS}_t$. 
The last 2-second accelerometer and gyroscope data are converted into the frequency domain using Power-Spectral Density (PSD) representation~\cite{youngworth2005overview} before being fed into the 2-layer inertia encoder and producing a 512-dimensional inertial representation $\psi^\textrm{I}_t$. All encoders are trained to minimize $\mathcal{L}_\textrm{TRON}$ (Eqn.~\eqref{eqn::cahsor_ssl}). 

\subsubsection{Kinodynamics}
Our $\mathbb{SE}(3)$ vehicle state is instantiated in the current robot frame, i.e., $[x_t, y_t, z_t, r_t, p_t, \phi_t]^T = \mathbf{0}$, and therefore omitted from the input of our forward kinodynamic model (Eqn. \ref{eqn::approximated_fkd}).
To explicitly showcase the efficacy of the learned kinodynamic model on state dimensions beyond $\mathbb{SE}(2)$, we limit the model output to three metrics to reflect sliding along $y$, roll around $x$, and bumpiness along $z$, i.e., $[\texttt{sliding}_{t+1}, \texttt{roll}_{t+1}, \texttt{bumpiness}_{t+1}]^T$. While it is not necessary for the human shared-control setting, for autonomous navigation planning, other state dimensions are produced using a simple Ackermann-steering model, whose predicted 3-DoF trajectories are evaluated for competence awareness with the learned kinodynamic model. 
Such a practice also avoids the computation overhead of sequentially rolling out a large set of multi-step, 6-DoF candidate trajectories, which cannot be efficiently parallelized on GPUs. Using a learned kinodynamic model for both competence awareness and trajectory rollout can be done with more onboard computation. 
To be specific, $\texttt{sliding}_{t+1}$ is captured by the ground speed sensed by GPS-RTK projected onto the robot $y$ axis (left);  
We compute the absolute angular acceleration around the $x$ axis (front) from the gyroscope averaged over 0.1s as $\texttt{roll}_{t+1}$; 
$\texttt{bumpiness}_{t+1}$ is computed as the absolute jerk along the $z$ axis (up) from the accelerometer averaged over 0.1s.
We find out such metrics can more precisely reflect undesirable vehicle $\mathbb{SE}(3)$ motions, compared to the raw and noisy GPS and IMU readings.
As a downstream task of \textsc{tron}, the kinodynamic model (Eqn. \ref{eqn::approximated_fkd}) is learned with three independent Multi-Layer Perceptron (MLP) heads. Each head consists of three hidden layers with dimensions of  256, 64, and 1 respectively. The MLPs take as input the pretrained visual, speed, and/or inertial representation $\psi^\textrm{V, S, I}_t$ and candidate control action $\mathbf{u}_t = (v, \omega)$ (omitted in Fig.~\ref{fig::nn} right for simplicity), to produce $\texttt{sliding}_{t+1}$, $\texttt{roll}_{t+1}$, and $\texttt{bumpiness}_{t+1}$. Notice that these metrics are simply three specific examples we choose to showcase $\mathbb{SE}(3)$ competence awareness, but other metrics can also be used when necessary. While manually-crafted rules can likely achieve similar results, they tend to be conservative due to the qualitative heuristics inherent in those rules. Furthermore, it is not scalable to design such rules or heuristics for every new aspect of $\mathbb{SE}(3)$ competence awareness of interest (e.g., constraining the pitch angle due to aggressive acceleration and deceleration) and new environmental factor (e.g., a new type of terrain). In contrast, \textsc{cahsor} provides both a quantitative and scalable way of enabling $\mathbb{SE}(3)$ competence awareness while precisely achieving the maximum possible speed.

\subsection{Autonomous Navigation Planning with \textsc{cahsor}}
We integrate our \textsc{cahsor} model with a Model Predictive Path Integral (\textsc{mppi}) planner~\cite{williams2017model}. Our \textsc{mppi} planner rolls out a set of candidate 3-DoF state trajectories using sampled action sequences and then combines those samples based on a predefined cost function.
In addition to minimizing the Euclidean distance to the goal and maximizing speed, the cost function is informed by the prediction of the learned 6-DoF kinodynamic model.
Furthermore, our model performs multi-step prediction in each \textsc{mppi} cycle with a planning horizon of 8 time steps and 550 action samples per cycle.
The model assigns infinitely large costs to candidate trajectories involving significant roll, sliding, and bumpiness.

\textsc{mppi} then updates the sampling distribution to sample actions that are more likely to lead to low cost trajectories, i.e., moving the robot toward a goal at the fastest possible speed.
We select six goals in an outdoor off-road environment for the robot to drive to in a loop (Fig.~\ref{fig::loop}). For \textsc{mppi} rollouts, future terrain inertial responses are not available to the \textsc{tron} model. Therefore, we only use the visual and speed representation $\psi^\textrm{VS}_t$ as $\psi^\textrm{V, S, I}_t$ to represent the world state $\mathbf{w}_t$ associated with each future vehicle state $\mathbf{s}_t$ on the candidate trajectories. For computation efficiency, we split the current BEV of 5.5m$\times$3m into a 15$\times$51 grid, resulting in patches of size 0.85m$\times$0.65m, which corresponds to the size of the robot footprint. We then pick the terrain patch that is closest to $\mathbf{s}_t$ on the candidate trajectories for parallelized model query on GPU during one \textsc{mppi} cycle.


\begin{table*}[t]
    \centering
    \begin{tabular}{ccccccc|cc|ccc}
        \toprule
        \textbf{Loss} &  \multicolumn{6}{c}{\textbf{\textsc{end-to-end}}}    & \multicolumn{2}{c}{\textbf{\textsc{sterling}}} & \multicolumn{3}{c}{\textbf{\textsc{tron}}}    \\ 
        & V & I & VI & VS & IS & VSI & V & VI & I & VS* & VSI* \\ 
        \midrule
        \textbf{Roll} & 0.32 & 0.10 & 0.07 & 0.23 & 0.11 & 0.09 & 0.27 & 0.05 & 0.03 & 0.08 & \textbf{0.02} \\
        \textbf{Sliding} & 0.69 & 0.61 & 0.54 & 0.46 & 0.45 & 0.42 & 0.62 & 0.47 & 0.44 & 0.35 & \textbf{0.29} \\
        \textbf{Bumpiness} & 0.24 & 0.05 & 0.04 & 0.17 & 0.06 & 0.05 & 0.17 & 0.04 & 0.02 & 0.06 & \textbf{0.02} \\
        \midrule
        \textbf{Combined} & 0.72 & 0.24 & 0.21 & 0.30 & 0.21 & 0.18 & 0.36 & 0.18 & 0.15 & 0.16 & \textbf{0.10} \\
        \bottomrule
    \end{tabular}
    \caption{Kinodynamics Test Loss with Different Representations of \texttt{vision} (V), \texttt{inertia} (I), and \texttt{speed} (S) using \textsc{end-to-end}, \textsc{sterling}, and \textsc{tron}. * denotes the models deployed on the physical robot.}
    \label{tab::ssl}
\end{table*}

\subsection{Human-Autonomy Shared-Control with \textsc{cahsor}}
We also demonstrate the use case of \textsc{cahsor} in a human-autonomy shared-control setup, in which a human driver aims at driving the robot as fast as possible, while \textsc{cahsor} takes care of satisfying all vehicle $\mathbb{SE}(3)$ constraints with the closet possible vehicle control to the human command. In this case, the objective function in Eqn.~\eqref{eqn::optimization} becomes 
\begin{equation}
        \mathbf{u}^*_t = \argmin_{\mathbf{u}_t} ||\mathbf{u}_t-\mathbf{u}^H_t||.
    \label{eqn::shared_autonomy}
\end{equation}
$\mathbf{u}^H_t$ is the desired human control input, which may violate the $\mathbb{SE}(3)$ constraints. 
In this shared-control setup, both inertia and vision (from past camera images) are available, so \textsc{tron} takes in visual, speed, and inertial representation as $\psi^\textrm{V, S, I}_t$ to represent the current world state $\mathbf{w}_t$.

\begin{figure*}
  \centering  \includegraphics[width=0.66\columnwidth]{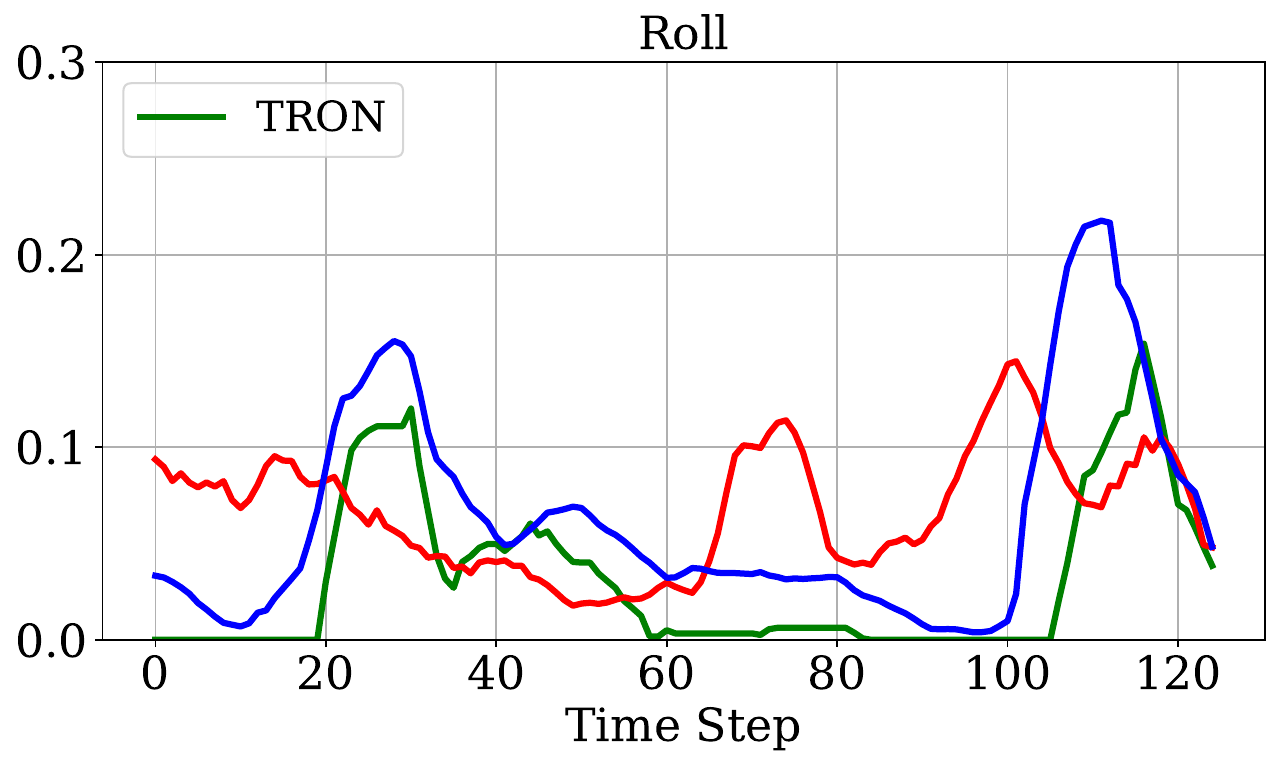}
  \includegraphics[width=0.66\columnwidth]{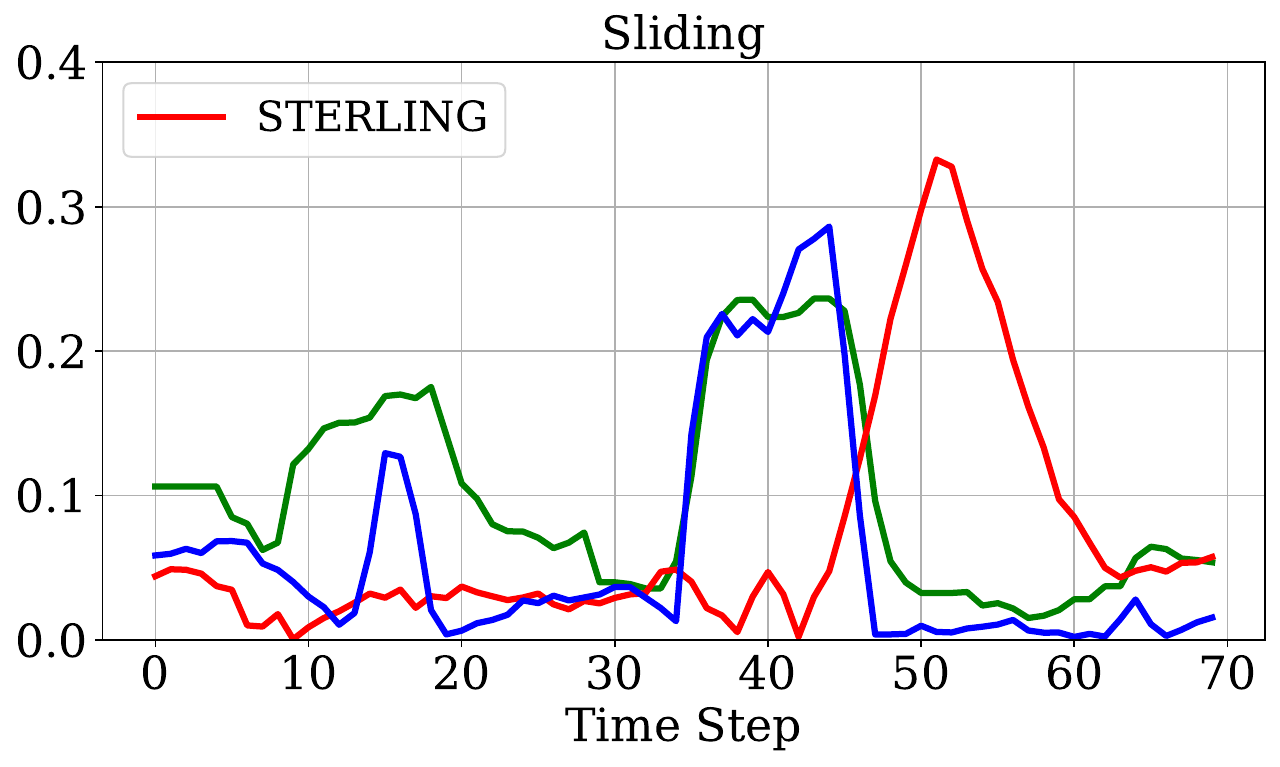}
  \includegraphics[width=0.66\columnwidth]{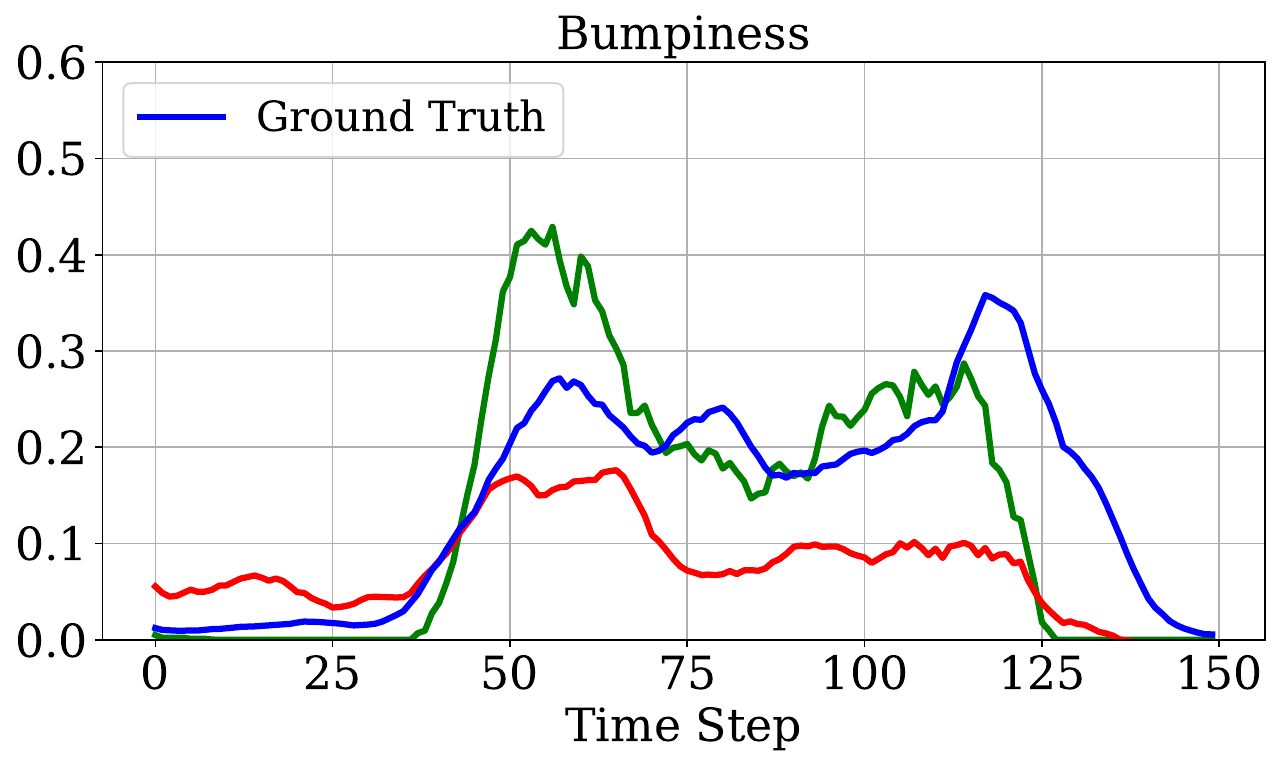}
  \caption{Downstream Kinodynamic Model Prediction with \textcolor{green}{\textsc{tron}} and \textcolor{red}{\textsc{sterling}} Pretraining Compared to \textcolor{blue}{Ground Truth}: \textsc{tron} produces both qualitatively and quantitatively accurate predictions, while \textsc{sterling} fails to qualitatively capture the changes in roll and sliding, but only gets the trend of bumpiness.}
  \label{fig::prediction_viz}
\end{figure*}

\section{EXPERIMENTS}
\label{sec::experiments}

We deploy \textsc{cahsor} navigation on the Hunter SE in locations different from where the training data is collected, but with similar terrain types. 
While the training data have been collected during early fall, due to the development phase, our final experiments are conducted in the winter. Therefore, the terrain features have experienced significant changes due to spatial and temporal differences, e.g., the grass has changed density, thickness, and color, the rocks have been displaced, and the pavement has been under different lighting conditions. Despite the generalization to such similar terrain with different characteristics, we do not expect our method to generalize to completely unseen terrain, e.g., mud or gravel, for which more training data may be necessary.

\subsection{\textsc{tron} Learning Results}
To demonstrate the effectiveness of the multimodal, i.e., visual, speed, and inertial, and self-supervised \textsc{tron} learning, we present \textsc{tron}'s downstream kinodynamics learning results compared against a set of baselines. To be specific, we implement \textsc{end-to-end} kinodynamics learning from \texttt{vision} (V), \texttt{inertia} (I), \texttt{vision \& inertia} (VI), \texttt{vision \& speed} (VS), \texttt{inertia \& speed} (IS), and all three (VSI), both as an ablation study for the proposed \textsc{tron} learning and to analyze the information contained in each (and different combinations of) perceptual modality. We also implement a representation learning approach to correlate only \texttt{vision} and \texttt{inertia} and deploy with \texttt{vision} only, without considering \texttt{speed}, similar to \textsc{sterling}~\cite{karnan2023sterling}. \textsc{tron} is experimented with \texttt{inertia}, \texttt{vision \& speed}, and \texttt{vision \& speed \& inertia} representation. Table \ref{tab::ssl} shows the test MSE loss on downstream kinodynamics learning tasks of predicting roll, sliding, bumpiness, and all three combined.

As \textsc{end-to-end} results show, \texttt{vision} contains the least amount of information for kinodynamic model learning, producing the highest losses on all dimensions. \texttt{inertia} contains much more information than \texttt{vision} and produces lower losses. Combining \texttt{vision} and \texttt{inertia} keeps reducing the losses. On the other hand, adding \texttt{speed} to \texttt{vision} significantly boosts performance, while \texttt{inertia} combined with \texttt{speed} only shows marginal improvement, possibly due to the fact that \texttt{inertia} already contains a large amount of information from \texttt{speed}. Using all three modalities results in the lowest combined loss overall for \textsc{end-to-end}. One important observation is that \texttt{inertia} significantly outperforms \texttt{vision \& speed} in most dimensions, except for sliding, in the \textsc{end-to-end} setup. 
However, \textsc{tron} is able to reduce such a performance gap by allowing them to augment each other during representation learning and maximize the information contained in their representation spaces, as they achieve comparably low losses. Combining all three after \textsc{tron} learning expectedly achieves the best results overall. 
\textsc{sterling} is designed to use \texttt{inertia} to augment \texttt{vision} so that information contained in \texttt{vision} can be maximized when only \texttt{vision} is available during deployment. While \textsc{sterling} works well on downstream terrain preference learning (e.g., grass is better than pebble)~\cite{karnan2023sterling}, not considering \texttt{speed} during representation learning introduces ambiguity in the representation spaces and therefore leads to bad performance on our kinodynamics learning task (e.g., what will happen when driving quickly/slowly on grass/pebble). For a fair comparison, we also show the results of \textsc{sterling} with both \texttt{vision} and \texttt{inertia} (which is not available for planning on future states). Despite improved performance compared to using \texttt{vision} alone, due to the missing \texttt{speed} information during representation learning, \textsc{sterling} still suffers from worse performance compared to \textsc{tron}. 

In Fig.~\ref{fig::prediction_viz}, we show the kinodynamic model prediction results using the pretraining from \textsc{tron} and \textsc{sterling} compared against ground truth values. The prediction from \textsc{tron} when considering both vision and speed (VS) can produce more similar roll, sliding, and bumpiness values to the ground truth, compared to \textsc{sterling} using only vision (V). Specifically, for roll and sliding, \textsc{tron} matches with ground truth both qualitatively (see the two peaks in both roll and sliding) and quantitatively (see the small difference between the green and blue lines), whereas \textsc{sterling} does not even qualitatively predict the trend correctly (red lines miss all blue peaks and randomly predict peaks at the wrong locations); for bumpiness, both \textsc{tron} and \textsc{sterling} qualitatively produce the correct trend, but \textsc{tron}'s prediction is quantitatively closer to the ground truth.

\subsection{Human-\textsc{cahsor} Shared Autonomy}
We conduct the human-\textsc{cahsor} shared autonomy experiments in three separate locations with different terrain types to provide a detailed view of how \textsc{cahsor} affects vehicle states in $\mathbb{SE}(3)$.
The fully manual teleoperation mode gives the human the full range of vehicle control commands, i.e., -4.8m/s to 4.8ms throttle and -0.5rad to 0.5rad steering.
\textsc{cahsor} selects actions based on Eqn.~\eqref{eqn::optimization} with the objective function in Eqn.~\eqref{eqn::shared_autonomy}. 
The human is asked to drive the robot through a rocky patch and maneuver aggressively on grass and pavement, all at the fastest speed. 

\begin{figure*}
  \centering
  \includegraphics[width=2\columnwidth]{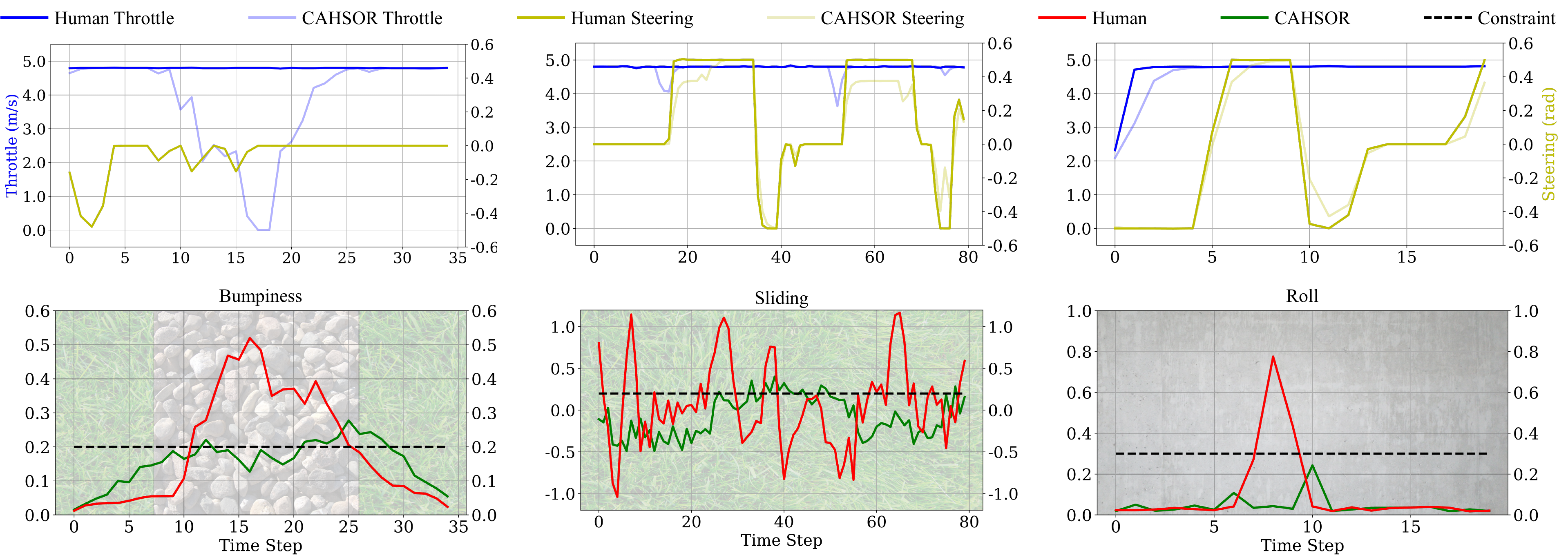}
  \caption{Examples of Human-\textsc{cahsor} Shared Autonomy on Rocks, Grass, and Pavement to Limit Bumpiness, Sliding, and Roll.}
  \label{fig::manual_exp}
\end{figure*}

We show the human commanded throttle and steering and the \textsc{cahsor} constrained commands in Fig.~\ref{fig::manual_exp} top, with resulted bumpiness, sliding, and roll (Fig.~\ref{fig::manual_exp} bottom green). For comparison, we drive the robot without \textsc{cahsor} and overlay the resulted bumpiness, sliding, and roll (Fig.~\ref{fig::manual_exp} bottom red) on the figure. \textsc{cahsor} knows to reduce speed on grass to avoid sliding and on pavement to prevent roll during high-speed turns, and to slow down on rocks to reduce bumpiness even without turning. 
However, due to the higher requirement of instantaneous accuracy of roll prediction compared to bumpiness and sliding, we observe that during on-robot experiments, \textsc{cahsor} is not able to prevent roll in advance all the time. 

\subsection{Autonomous \textsc{cahsor} Navigation}
The autonomous \textsc{cahsor} navigation using \textsc{mppi} is compared against five baselines: (1) vanilla \textsc{mppi} without \textsc{cahsor} with the same maximum speed of 4.8m/s; (2) the 4.8m/s \textsc{mppi} with \textsc{sterling}~\cite{karnan2023sterling}, a state-of-the-art off-road representation learning method which only considers the visual, but not speed representation (while inertial information is not available for future states); (3) a recent high-speed kinodynamic modeling approach used with \textsc{mppi} by~\citet{lee2023learning}; (4) a slow \textsc{mppi} with a maximum speed of 3.0m/s, which aims to demonstrate that conventional approaches need to slow down much more than \textsc{cahsor} in order to achieve similar vehicle stability in $\mathbb{SE}(3)$; and (5) the slow 3.0m/s \textsc{mppi} with \textsc{cahsor} to show that \textsc{cahsor} does not jeopardize navigation performance by unnecessarily reducing speed.

\begin{table*}
    \centering
    \begin{tabular}{cccccc}
        \toprule
         & \begin{tabular}{c}Top \\Speed (m/s)\end{tabular}~\small$\uparrow$ & \begin{tabular}{c}Average \\ Speed (m/s)\end{tabular}~\small$\uparrow$ & \begin{tabular}{c}Average \\\texttt{bumpiness}\end{tabular}~\small$\downarrow$ &\begin{tabular}{c}Maximum \\ \texttt{sliding}\end{tabular}~\small$\downarrow$ &\begin{tabular}{c} Maximum \\ \texttt{roll}\end{tabular}~\small$\downarrow$ \\ 
        \midrule
        4.8m/s Autonomous+\textsc{cahsor} & 4.60 & 3.92$\pm$0.12 & \textbf{0.054$\pm$0.007} &  \textbf{0.62} & \textbf{0.11}\\
        4.8m/s Autonomous & \textbf{4.80} & \textbf{4.29$\pm$0.10} & 0.110$\pm$0.007 & 1.11 & 1.20 \\
        4.8m/s Autonomous+\textsc{sterling}~\cite{karnan2023sterling}  & 4.75 &  3.98$\pm$0.08  & 0.075$\pm$0.004 & 0.83 & 0.98\\
        4.8m/s Autonomous+\citet{lee2023learning} Model & 4.72 &  4.23$\pm$0.05  & 0.106$\pm$0.003 & 0.91 & 1.17\\
        \midrule
        3.0m/s Autonomous & 2.75 &  2.45$\pm$0.02  & 0.065$\pm$0.010 & 0.69 & 0.12\\
        3.0m/s Autonomous+\textsc{cahsor} & 2.73 &  2.38$\pm$0.07  & 0.052$\pm$0.003 & 0.60 & 0.05\\
        \bottomrule
    \end{tabular}
    \caption{Speed and $\mathbb{SE}(3)$ Competence Awareness Achieved by Six Different Methods on Unseen Terrain.}
    \label{tab::loop}
\end{table*}

\begin{figure}
  \centering
  \includegraphics[width=1\columnwidth]{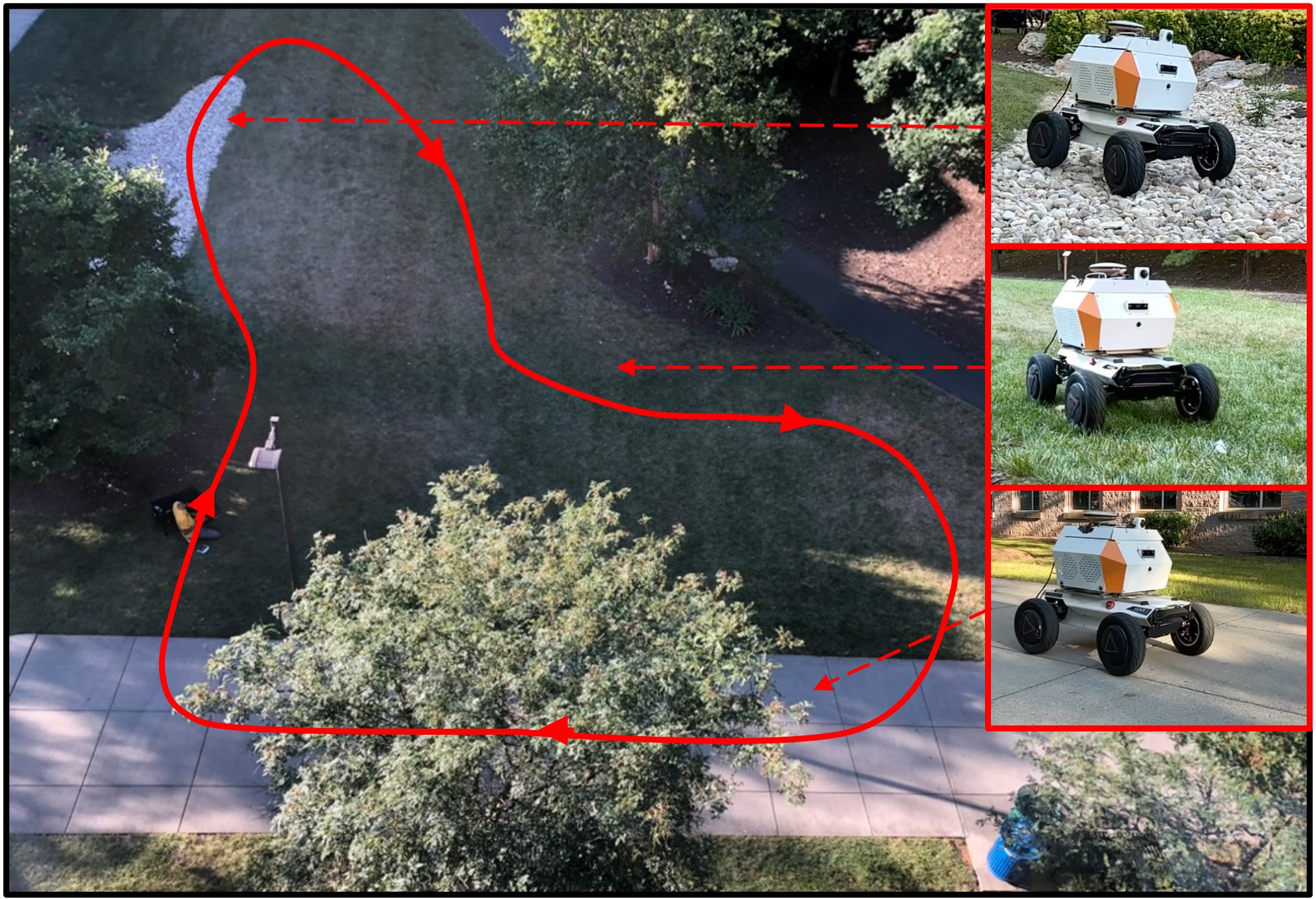}
  \caption{Autonomous Navigation in a Loop (Three Loops each Method).}
  \label{fig::loop}
\end{figure}

Table \ref{tab::loop} shows the top speed, average speed, average \texttt{bumpiness}, maximum \texttt{sliding}, and maximum \texttt{roll} achieved by the six approaches. 4.8m/s autonomous navigation reaches the highest top speed of 4.8m/s and achieves an average of 4.29m/s around all three loops, only decelerating to reach each pre-defined GPS waypoint. However, it experiences significant bumpiness, sliding, and roll motions along the way. The high bumpiness may damage onboard components. In fact, we have to stop the experiments twice in order to fix a loose USB connection due to extensive vibration on the rocks (Fig.~\ref{fig::loop} top left and Fig.~\ref{fig::hunter} lower left). Large roll angles also appear many times when executing the sharp turn from cement pavement to grass (Fig.~\ref{fig::loop} lower left and Fig.~\ref{fig::hunter} top). 
When turning around obstacles, excessive sliding that is not considered by normal $\mathbb{SE}(2)$ models may risk collision (Fig.~\ref{fig::loop} middle and Fig.~\ref{fig::hunter} lower right). On the other hand, autonomous navigation assisted by \textsc{cahsor} achieves a 4.6m/s top speed and slows down more, not only in order to reach GPS waypoints, but also due to competence awareness. We observe the vehicle significantly slows down when about to enter the rock patch to reduce bumpiness and executes less sharp or slower turns on grass and on cement to avoid too much sliding and roll. 
Assisted with visual representation alone (V), \textsc{sterling} reduces the top and average speed of the 4.8m/s navigation system to 4.75m/s and 3.98m/s respectively so as to improve the competence-awareness metrics, but not as much as \textsc{cahsor}. The reason is due to the lack of consideration of speed (S) compared to \textsc{cahsor}, whose VS representation has been correlated with inertia (I) during \textsc{tron} pretraining. The model by \citet{lee2023learning} considers the underlying elevation map due to undulating terrain and has shown promising results in simulation with high-precision vehicle state information. Since our off-road environment is relatively flat, for a fair comparison, we replace the elevation map with the visual terrain patch to incorporate semantic information. Without \textsc{tron} pretraining, the model by \citet{lee2023learning} achieves worse competence awareness compared to \textsc{sterling}, especially almost causing vehicle rollover.
The slow 3.0m/s autonomous navigation approach achieves a top speed of 2.75m/s and an average speed of 2.45m/s. This slow autonomous navigation system clocks much longer lap time compared to its full speed counterpart without or with \textsc{cahsor}. Even with a low speed, the slow system still cannot achieve better average \texttt{bumpiness}, maximum \texttt{sliding}, and maximum \texttt{roll} as 4.8m/s Autonomous+\textsc{cahsor} does, because it does not know when it is necessary to slow down to reduce such undesired movement and when it is possible to accelerate to achieve high speeds. 
Lastly, the slow 3.0m/s system with \textsc{cahsor} only marginally reduces top and average speed with marginally improved vehicle stability metrics, which shows that \textsc{cahsor} does not unnecessarily slow down the vehicle to blindly improve competence-awareness, risking jeopardizing the performance of a conservative planner that will not cause catastrophic consequences in the first place.
The GPS waypoint loop experiments show that \textsc{cahsor} is able to maintain a very high average speed and only slows down when the $\mathbb{SE}(3)$ constraints would be violated.

\section{CONCLUSIONS} 
\label{sec::conclusions}

Our \textsc{cahsor} ground navigation approach is able to utilize multimodal, self-supervised terrain representation, i.e., \textsc{tron}, to reason about the consequences of taking aggressive maneuvers on different off-road terrain, i.e., being competence-aware. Inertial observations contain the most information to enable efficient kinodynamics learning, but may not be available during planning. Augmenting easily available vision combined with speed using inertia with \textsc{tron}, similar kinodynamics learning performance can be achieved. Extensive physical experiments in both an autonomous navigation planning and human shared-control setup demonstrate \textsc{cahsor}'s superior competence awareness during high-speed off-road navigation. 

\section*{ACKNOWLEDGEMENT}

This work has taken place in the RobotiXX Laboratory at George Mason University. RobotiXX research is supported by National Science Foundation (NSF, 2350352), Army Research Office (ARO, W911NF2220242, W911NF2320004, W911NF2420027), US Air Forces Central (AFCENT), Google DeepMind (GDM), Clearpath Robotics, and Raytheon Technologies (RTX).



\bibliographystyle{plainnat}
\bibliography{mybib}

\end{document}